# Structural Combinatorial of Network Information System of Systems based on Evolutionary Optimization Method

Tingting Zhang†
Computer and military software engineering
PLA Army Engineering University
Nanjing China
101101964@seu.edu.cn

Yushi Lan
National defense key laboratory
China Electronics Technology Group Corporation
Nanjing, China
lan_ys@126.com

Aiguo Song
college of instrument science and engineering
Southeast University
Nanjing China
a.g.song@seu.edu.cn

Kun Liu
College of Command and Control Engineering
PLA Army Engineering University
Nanjing, China
15906183233@139.com

Nan Wang
College of Command and Control Engineering
PLA Army Engineering University
Nanjing, China
13851906714@139.com

## ABSTRACT

The network information system is a military information network system with evolution characteristics. Evolution is a process of replacement between disorder and order, chaos and equilibrium. Given that the concept of evolution originates from biological systems, in this article, the evolution of network information architecture is analyzed by genetic algorithms, and the network information architecture is represented by chromosomes. Besides, the genetic algorithm is also applied to find the optimal chromosome in the architecture space. The evolutionary simulation is used to predict the optimal scheme of the network information architecture and provide a reference for system construction.

## CCS CONCEPTS

• Computer systems organization ~ Architectures ~ Distributed architectures ~ Grid computing

**ACM Reference format:**

†Corresponding Author: Tingting Zhang (1978.05-), female (Chinese), associate professor, Ph.D., the main research areas are software engineering, systems engineering, and intelligent control systems (101101964@seu.edu.cn, zhangtings@sohu.com, Qinhuai, Nanjing Guanghua Road, Haifu Lane, School of Command and Control Engineering, Army Engineering University)

Tingting Zhang, Yushi Lan, Aiguo Song, Kun Liu, Nan Wang. 2020. Structural Combinatorial of Network Information System of Systems Based on Evolutionary Optimization Method. In *KDD'20: The 26th ACM SIGKDD International Conference on Knowledge Discovery Data Mining, August 22-27, 2020, California, USA, 8 pages*.

## 1  Preface

For the planning and construction of the network information system, the main concern is to meet the specific mission tasks of the final capability of the network information system. In the system planning stage, combing with experience, the system builder selects the member systems and interface set of the system, and all the architecture plans to form the initial architecture space according to the existing system conditions, technical force, and the resources that can be allocated. The system builder expects to evaluate the system construction scheme before the system is built, predict the results of the system evolution, and provide system architecture reference basis. If all the system architectures are subjected to evolutionary evaluation, a large volume of work is expected. Therefore, optimizing and evaluating the optimal system architecture before evaluating the quality of it can save much time for the sake of the whole construction.

In view of the above problems, this paper proposes a multi-objective meta-architecture optimization selection method based on genetic algorithms. Firstly, the structure of network information architecture space is described: the chromosome set is used to represent the network information architecture space. Then, a genetic algorithm-based chromosome set optimization strategy is proposed, which includes defining parameters, building models, and algorithm implementations.

## 2  Description of network information architecture space

The capability baseline constituting the system is set according to the mission. In order to offer services, the primary system sends a request for participation to sub-systems, and the sub-system may agree or reject for the sake of its own interests. In order to ensure the basic capability composition, the system sends requests to multiple systems with the same capabilities so that multiple systems can be used as alternatives. Meta-architecture is all the architectures that make up the system's capabilities. The architecture alternative space that makes up the initial system's capabilities chooses the optimal structure and simulates its evolution. Assume that for Network Information System Capability $SoS.C$, which contains the capabilities $C_i = (C_1, C_2 \ldots C_n)$.

Assume that for $\forall C_i \in SoS.C$, it presences $N_i$ member system combinations that satisfy its system capability $C_i$, namely $C_i$; it presences $\prod_{i=1}^{n} N_i$ system's dynamic integration schemes making up the architecture space $SoS.A_0$ [1]. However, it is a huge workload to perform the evolutionary simulation analysis of each of the alternative architectures in the system, which may lead to the "explosion" dilemma of the workload. In order to solve this problem, an effective optimization strategy needs to be used to analyze each dynamic integration scheme, find the optimal (better) member system dynamic integration scheme, and then perform evolutionary simulation of the architecture through evolutionary strategies.

## 3 Chromosome representation of network information architecture

We represent all the system architectures that make up the initial capabilities of the network information system by genome $SoS.A_o$ [2]. The specific method is: each system architecture is represented by a chromosome and in each chromosome $S_i (i = 1, 2 \ldots n)$ indicates the system, $I_{ij}$ represents the interface between system $S_i$ and the system $S_j$. If the chromosomes $S_i$ displays "1", the system $S_i$ will participate in and support SoS capabilities and $SoS.C$ construction; If the chromosome $S_i$ displays "0", the system $S_i$ will not participate in and support SoS capabilities and $SoS.C$ construction. If the chromosome $I_{ij}$ displays "1", there is an interconnection between the system $S_i$ and the system $S_j$; If the chromosomes $I_{ij}$ displays "0", there is no interconnection between the system $S_i$ and the system $S_j$. A chromosome describes whether the systems and interface that constitute the architecture work or not. Therefore, a chromosome describes a possible architecture, and all chromosomes constitute the initial architectural space. The chromosome representation architecture of network information system architecture $SoS.A_o$ is shown in Table 1:

**Table 1: Chromosome representation network information architecture**

| $S_1$ | $S_2$ | ... | $S_i$ | ... | $S_n$ | $I_{12}$ | ... | $I_{1n}$ | ... | $I_{ij}$ | ... | $I_{n-1,n}$ |
|---|---|---|---|---|---|---|---|---|---|---|---|---|
| 1 | 0 | | 1 | | 1 | 1 | | 0 | | 1 | | 0 |

For achieving system capabilities $SoS.C$, there are $M_i$ alternative systems, and $L_i$ possible interfaces. Each code represents one choice of architecture, then $X_i$ indicates $X_i$ interfaces of $N_i$ systems in $SoS.A_o$ ($1 \leq X_i \leq N_i$)。

## 4 Architecture multi-objective optimization strategy

The evaluation of the architecture is a vague problem, because the evaluation criteria are often non-quantitative and non-objective, which should be based on the unknown future situation, such as the "robustness" indicator. As the composition of the system cannot be predicted in advance, and the system participation is binary, the genetic algorithm can be applied to the study of the architecture as a non-gradient optimization algorithm. The genetic algorithm makes full use of the characteristics of the chromosome based on the regional changes of the system to plan the member system. At the same time, an objective optimization function will be used to determine the appropriate evolution direction of the chromosome.

### 4.1 Basic idea of genetic algorithm

Based on the ideas of biological evolution, genetic algorithm is a self-adapted global optimization probability search algorithm formed by simulating the genetic mechanism and natural selection organisms [3,4,5]. The genetic algorithm regards the optimization problem as a "chromosome" and measures the pros and cons of the "chromosome" through the fitness function according to the "Survival of the Fittest". The "chromosome" with the large fitness value has a high probability of being retained in genetic operations in the process of replication, crossover, and mutation. After multiple generations of evolution, the algorithm converges to the "chromosome" with the best fitness value, which is the optimal solution or the better solution of the optimization problem.

In the genetic algorithm, the population size $N$ of each generation is fixed. It has a certain number of iterations (150) for the termination of the algorithm, and the chromosome with the largest fitness value is the optimal solution for the optimal selection problem. The basic process of the genetic algorithm is as follows:

1. Initialize the control parameters, set the size of the group to $N$, and the probability of crossing $P_c$, mutation probability $P_m$, termination rules (number of iterations $Gen = 150$).
2. The algorithm starts to execute: the number of iterations $Gen = 0$, and the number of individuals in the population $i = 0$.
3. Under a certain coding strategy, an initial population consisting of $N$ initial individuals is randomly generated.
4. Determine whether the termination rule of the genetic algorithm is satisfied. If it is satisfied, the search ends; otherwise, the following steps are performed.
5. Calculate each individual in the population according to the fitness function $x_i (i = 1, 2, \ldots, N)$ fit value $F(x_i)$;
6. According to the individual adaptation value $F(x_i)$ $(i = 1, 2, \ldots, N)$, calculate the individual's selection

probability $P(x_i) = \frac{F(x_i)}{\sum_{j=1}^{N} F(x_j)}$. Randomly select N individuals from the population to obtain the population;

7. Based on cross probability $P_c$, Select two individuals from the population, cross them, form two new offspring individuals, and add them to the new population; while the individuals that have not crossed in the population are directly copied to the new population;
8. Based on mutation probability $P_m$, Select individuals in the new population for a genetic mutation, and the new individuals after the mutation replace the individuals in the new population;
9. Algebra of iteration: $Gen = Gen + 1$ Go to step 4.

Aiming at the dilemma of the above algorithm "explosion", the genetic algorithm was improved and designed, and a genetic algorithm with constraint rules was implemented. The network information system dynamic integration scheme optimization selection model was given to make it suitable for the network information system dynamic integration scheme optimization selection.

### 4.2 Optimal model

An optimal model is defined as follows:

$$MaximizeP_{SoS} = \max_{\forall i}(P_i)$$

$$MinimizeF_{SoS} = \sum_{i=1}^{n} F_i$$

$$MinimizeD_{SoS} = \max(D_i)$$

$D_i$ is the time period to realize the capability $C_i$.

The goals are:

$$\sum_{i=1}^{n}\sum_{j=1}^{m} f_{ij}S_{ij} + \sum_{k=1}^{l} f_{ij}c_{ik} \leq F_i \quad \forall i$$

$$\max_{\forall j}(d_{ij}S_{ij}) \leq D_i \quad \forall i$$

$$\max_{\forall k}(d_{ik}c_{ik}) \leq D_i \quad \forall i$$

$$\max_{\forall j}(P_{ij}S_{ij}) \geq P_i \quad \forall i$$

$$\max_{\forall k}(P_{ik}c_{ik}) \geq P_i \quad \forall i$$

$$\sum_{j=1}^{m} S_{ij} \geq 1 \quad \forall i$$

$$S_j = \begin{cases} 1 & \sum_{i=1}^{n} S_{ij} \geq 1 \\ 0 & \sum_{i=1}^{n} S_{ij} = 0 \end{cases} \forall j$$

$$I_k = \begin{cases} 1 & \sum_{i=1}^{n} c_{ik} \geq 1 \\ 0 & \sum_{i=1}^{n} c_{ij} = 0 \end{cases} \forall k$$

$$I_k = \begin{cases} 1 & S_j = 1 \,\&\&\, S_{j'} = 1 \\ 0 & S_j = 0 \,||\, S_{j'} = 0 \end{cases}$$

$I_k$ is the interface of $S_j$ and $S_{j'}$, and $j \neq j'$.

### 4.3 Algorithm implementation

*4.3.1 Encoding.* Coding is the representation of one solution of the system dynamic integration solution optimization selection problem in a code manner so that the decoding space corresponds to the genetic code space of the genetic algorithm. In the solution, various combinations of member system $S_i$ and interface $I_{ij}$ constitute an architectural space $SoS.A_o$. Each of these architectures can achieve the initial capabilities of the network information system.

*4.3.2 Generation of the initial population.* In genetic algorithms, there are usually two methods for generating the initial population: one is the method of generating a random initial population, which is suitable for cases where there is no prior knowledge of the solution to the problem; the other is based on certain prior knowledge and transforms the prior knowledge into a set of constraints that must be satisfied, and then randomly selects the population's chromosomes based on these constraints.

Among the two initial population generation methods above, the second method is obviously easier than the first one in improving the efficiency of the genetic algorithm and obtain the optimal solution faster. We also use the initial population generation method in this paper. A method with constraint rules is used to obtain the initial population. The algorithm flow of its initial population generation is as follows:

1. Initialize, set population $SoS.A_o$ of size $popsize$.
2. Start execution, the number of chromosomes in the initial population $ChromNum = 0$.
3. IF $ChromNum = popsize$, The algorithm ends and the initial population is output; otherwise, the following steps are performed.
4. Let the number of chromosomes be $ChromNum = ChromNum + 1$, The $ChromNum$ th chromosome in the population is $SoS[ChromNum]$, and initialize the number of genes in the chromosome $geneNum = 0$.
5. IF $geneNum = n$, go to step 3, otherwise, perform the following steps;
//Determine whether the number of genes in the chromosome is satisfied $n$, $n$ is the number of systems.
6. Let the number of genes in the chromosome be $geneNum = geneNum + 1$. Get $n_{geneNum}$ value.
//$n_{geneNum}$Express The number of systems to choose from in $SoS_{geneNum}$.
7. $CR = getConflictRule(TA_{geneNum})$.
//Get $SoS_{geneNum}$ Preorder structure $SoS.A_1 ..., SoS.A_{geneNum-1}$ Constraint set $CR$.
8. $X_{geneNum} = CreatGene(SoS.A_{geneNum}, CR)$.

    // Generate $TA_{geneNum}$ Gene according to constraint rules.
9. $setConflictRule(X_{geneNum}, SoS.A, cr)$.
    //Generate post-order structures $cr$ according to $SoS.A_{geneNum}$ of $x_{geneNum}$ Constraints.
10. Go to step 5.

*4.3.3 Adapted value function design* In genetic algorithms, the main function of the fitness function is to comprehensively evaluate the pros and cons of the chromosome (the solution of the problem) to the environment so that the evolution of the chromosome can be optimized. The construction of the fitness function directly affects the convergence speed and search results of the genetic algorithm.

According to the analysis of network information system construction requirements, the objective functions selected for the optimization of the architecture are:

Objective function 1, $F_1$: The overall system performance is optimal ($P_{SoS}$)

$$Maximize P_{SoS} = \max_{\forall i}(P_i)$$

Objective function 2, $F_2$: Least budget for system construction ($F_{SoS}$)

$$Minimize F_{SoS} = \sum_{i=1}^{n} F_i$$

Objective function 3, $F_3$: The shortest time to achieve system functions ($D_{SoS}$)

$$Minimize D_{SoS} = \max(D_i)$$

Goal constraints:

$$\sum_{i=1}^{n}\sum_{j=1}^{m} f_{ij}S_{ij} + \sum_{k=1}^{l} f_{ij}c_{ik} \leq F_i \quad \forall i$$

$$\max_{\forall j}(d_{ij}S_{ij}) \leq D_i \quad \forall i$$

$$\max_{\forall k}(d_{ik}c_{ik}) \leq D_i \quad \forall i$$

$$\max_{\forall j}(P_{ij}S_{ij}) \geq P_i \quad \forall i$$

$$\max_{\forall k}(P_{ik}c_{ik}) \geq P_i \quad \forall i$$

The above three objective functions have obstacles to form a consistent adaptation value function. Therefore, the objective function needs to be normalized accordingly.

Firstly, transform the objective function 2 ($F_2$) into a maximized objective function:

$$F_2' = \max(\frac{1}{F_{SoS}})$$

Secondly, transform the objective function 3 ($F_3$) into an objective function that minimizes the time required to build the architecture ($F_3'$), then:

$$F_3' = \max(\frac{D_{SoS} - \min T}{D_{SoS}})$$

$D_{SoS}$ is the expected time of system construction.

Architecture collection $b$ determines the optimal performance architecture $a$:

$$P_{SoS,ab} = \max_{\forall i}[\max_{\forall j}(p_{ij}S_{ij}), \max_{\forall k}(p_{ik}c_{ik})] \mid ab$$

Architecture collection $b$ determines to achieve the least cost architecture $a$:

$$F_{SoS,ab} = \sum_{i=1}^{m}(\sum_{j=1}^{m} f_{ij}S_{ij} + \sum_{k=1}^{l} f_{ik}L_{ik}) \mid ab$$

Architecture collection $b$ determines minimum construction time architecture $a$:

$$D_{SoS,ab} = \max_{\forall i}[\max_{\forall j}(d_{ij}S_{ij}), \max_{\forall k}(d_{ik}c_{ik})] \mid ab$$

After the conversion of the objective function, all three objective functions achieve the maximum goal. As decision-makers have different preferences for each target, it is necessary to divide target weights. However, it is difficult to accurately define target weights generally in advance as the value of target weights may be different due to changes in the combat environment and specific combat tasks. So for specific combat missions, target weights change dynamically.

Starting from the general realization of the architecture of combat missions, expert evaluation methods, analytic hierarchy processes, and other quantitative target weight intervals can be used to provide general weight guidance for system construction command decision-makers or decision-making agencies. Suppose that the objective function is $F_1'F_2'F_3'$ and the weight intervals are $[w_a^1, w_b^1], [w_a^2, w_b^2], [w_a^3, w_b^3]$ There is a relationship between the weight intervals: (1) $w_a^1 + w_a^2 + w_a^3 < 1$, （2）$w_b^1 + w_b^2 + w_b^3 > 1$.

Starting from the realization of specific combat tasks, architecture construction decision-makers extract weight values $w_p$、$w_f$、$w_d$ from the above three intervals according to their own knowledge and experience and finally get the fitness value function of the genetic algorithm $F$:

$$F = \varphi_{SOS,ab} = w_p P_{SOS,ab} + w_d D_{SOS,ab}$$

*4.3.4 Select operation.* The first operation of the selection operation is to calculate the adaptation value of each chromosome in the population according to the adaptation value function. The main purpose of this operation is to make a better chromosome survive as much as possible. However, it is not possible to sort and select based on the chromosome adaptation value alone, because this will easily cause diversification to disappear and lead to precocity. Therefore, in the selection operation, each chromosome is given a selection probability in the selected population, so that all chromosomes have a certain chance of survival. Of course, the stronger the ability to adapt to the environment, the greater the probability of survival will be. In particular, it is emphasized that in combination with the elite selection strategy, the chromosome with the largest adaptation value in the population is always copied to the next generation population.

The algorithm flow of the selection operation is divided into three stages:

Phase 1: The calculation of the single chromosome adaptation value and the sum of all chromosome adaptation values in the population. The algorithm flow is as follows:

1. Initialize and get the previous generation population $POP$ and the size of its population is $popsize$.
2. Fitness value calculation. Set the initial chromosome fit value $CFitness$ Number of calculations $CFNum = 0$, the sum of chromosome adaptation values $SumCF = 0$;
3. IF $CFNum = popsize$, computation ends and output the adaptation values of all chromosomes and the adaptation values of the population; otherwise, perform the following steps.
4. Set $CFNum = CFNum + 1$. Calculation $F(POP[CFNUM])$.
5. $CFitness[CFNum] = F(POP[CFNum])$ , $SumCF = SumCF + CFi$.
6. Go to step 3.

Phase 2: Single chromosome survival probability $P_i$ calculation. The algorithm flow is as follows:

1. IF $CFNum = popsize$, The calculation ends. Output the survival probability of all chromosomes; otherwise, the following steps are performed.
2. The survival probability of the $CFNum$ bar staining is: $P_{CFNum} = \frac{CFitness[CFNum]}{SumCF}$.
3. Go to step 1.

Phase 3: According to the survival probability of chromosomes, the roulette method is used to select chromosomes. The algorithm flow is as follows:

1. Initialize, sort all chromosomes according to the survival probability, and obtain the chromosome with the highest survival probability. Copy it directly to the next generation population, and then calculate the cumulative probability for the remaining chromosomes to obtain the survival interval of each chromosome $[a_i, b_i]$, which satisfies:

$$P_i = b_i - a_i$$

$$a_i = \sum_{j=1}^{i-1} P_j$$

2. $getMaxCFitness(POP)$.

   //Get the chromosome with the highest applicability value in the previous population.

3. Set the number of selected chromosomes $ChoiceNum = 1$.
4. IF $ChoiceNum = popsize - 1$, select the end and output the next generation population. Otherwise, perform the following steps.
5. In the interval $[0,1]$, generate a uniformly distributed random number $x = Random(0,1)$.

   IF $(a_i < x \leq b_i)$

   Keep the No. $i$ Chromosome of the first generation of the population.

   $ChoiceNum = ChoiceNum + 1$.

   End IF

6. Go to step 4.

*4.3.5 Cross operation.* Suppose the crossover probability of the poor individual is $P_{c1}$, the crossover probability of the optimal individual is $P_{c2}$. Calculate the probability of crossover of chromosomes $P_c$ for:

$$P_c = \begin{cases} P_{c1} & f \leq \overline{f} \\ P_{c1} - \frac{(p_{c1} - p_{c2})(f - \overline{f})}{f_{max} - \overline{f}} & f > \overline{f} \end{cases} \quad (1)$$

In the formula (1), $f$ indicates the chromosome with the larger adaptation value on both sides of the crossover operation, $\overline{f}$ is the average fit value of the population, and $f_{max}$ is the largest fitness value in the population.

Besides, on the cross rules, a change cross method is adopted to ensure the effectiveness of the cross operation. The flow of the change cross algorithm is as follows:

1. Initialize and obtain two parent chromosomes $POP[i]$, $POP[j]$ for cross over the operation, and their adaptation values in the population are $CFitness[i]$、$CFitness[j]$.
2. Calculate the crossover probability $P_c$ of the two selected chromosomes according to formula (1).
3. In the interval $[0,1]$, generate a uniformly distributed random number $c = Random(0,1)$.

   IF $P_c < c$

   // Calculate the set of genetic differences between two chromosomes.

   $$D = \{d_1, d_2, ..., d_m\}.$$

For $\forall dk \in D$, there must exist $x$ that $dk = \sum_{i=1}^{x} N_i$.

Perform step 4

Else

Perform step 1

End IF

4. When IF $D = \varnothing$, save the chromosome directly to the next generation; otherwise, perform the following steps.
5. In the interval $[0,1]$, generate a uniformly distributed random number $y = Random(0,1)$.
6. When it satisfies $\frac{a}{m} \leq y < \frac{a+1}{m}$, the intersection of chromosomes is $da = \sum_{i=1}^{q} N_i$, implement cross operations.
7. $CRAll = getAllConflictRule(TA_{da})$;

   // Get the architecture and all constraints related to $TA_{da}$;

8. $isMeet = JudgeConflictRule(new\_POP[i], new\_POP[j])$.
9. IF($isMeet$) copies $new\_POP[i]$ and $new\_POP[j]$ to the next generation; otherwise, keep the parent.
10. $GetTwoMaxFitNess(POP[i], POP[j], new\_POP[i], new\_POP[j])$.

    //Choose two chromosomes in the two generations with high fitness values of the father and son to avoid degradation in the genetic algorithm.

*4.3.6 Mutation operation.* Assuming the maximum probability of variation is $P_{m1}$, the minimum probability of variation is $P_{m2}$, and the adaptive mutation probability $P_m$. The calculation expression is:

$$P_m = \begin{cases} P_{m1} & f \leq \overline{f} \\ P_{m1} - \dfrac{(p_{m1} - p_{m2})(f_{max} - f)}{f_{max} - f} & f > \overline{f} \end{cases} \quad (2)$$

In formula (2), $f$ represents the fitness value of the chromosome to be mutated, $\overline{f}$ is the average fit value of the population, and $f_{max}$ is the largest fitness value in the population.

The process of implementing mutation operation algorithm is as follows:

1. Initialize, calculate the probability $P_m$ of mutation of each chromosome in the population according to formula (2).
2. In selection operation, use the roulette method to randomly select a chromosome according to the probability of mutation, and set $POP[i]$.
3. Generate a mutation position $i$ randomly. Call the constraint rule for this gene position $CRAll = getAllConflictRule(TA_{da})$ to regenerate a random number and transform this gene into a new chromosome $POP[new]$.
4. Calculate the adaptation value of the new chromosome $CFitness[new]$.

Assignment architecture collection $\overrightarrow{A_{SoS,b}}$. The best suitability value in set $b$ is:

$$\overline{\varphi_{SoS,ab}} = \max_a (\varphi_{SoS,ab})$$

The most applicable value for the entire architecture space is

$$\overline{\overline{\varphi_{SoS,ab}}} = \max_b (\overline{\varphi_{SoS,ab}})$$

Assign an architecture in the architecture set $b$, which is the architecture of the system and interfaces that provide all system capabilities.

$$A_{SoS,ab} = \sum_{i=1}^{n} A_{iab} \left| \exists \left\{ S_j \in A_{SoS,ab} = \begin{cases} 1 & \text{if } \sum_{i=1}^{n} S_{ij} \in A_i \geq 1 \\ 0 & \text{others} \end{cases} \forall j \right\} \right.$$

$$\cup \left\{ I_k \in A_{SoS,ab} = \begin{cases} 1 & \text{if } \sum_{i=1}^{n} c_{ik} \in A_i \geq 1 \\ 0 & \text{others} \end{cases} \forall k \right\}$$

Determine the best architecture $\overline{A_{SoS}}, \overline{ab}$ of the best applicability in the architectural set $b$ by setting the respective architecture systems and interfaces of all capabilities.

$$\overline{A_{SoS,\overline{ab}}} = \sum_{i=1}^{n} A_{i\overline{ab}} \left| \exists \left\{ S_j \in \overline{A_{SoS,\overline{ab}}} = \begin{cases} 1 & \text{if } \sum_{i=1}^{n} S_{ij} \in A_i \geq 1 \\ 0 & \text{others} \end{cases} \forall j \right\} \right.$$

$$\cup \left\{ I_k \in \overline{A_{SoS,\overline{ab}}} = \begin{cases} 1 & \text{if } \sum_{i=1}^{n} c_{ik} \in A_i \geq 1 \\ 0 & \text{others} \end{cases} \forall k \right\}$$

$\overline{\overline{A_{SoS}}}$ is the final architecture space the optimal architecture with an optimum suitability ($\overline{\overline{\varphi_{SoS}}}$).

5. IF $CFitness[new] \geq CFitness[i]$. When it is true, new chromosomes are added to the new population; otherwise the original chromosomes are saved;
6. End.

*4.3.7 Algorithm termination.* The genetic algorithm does not know when to end the search by itself, so we need to set certain termination rules. Common termination rules include the number of evolutionary generations, the calculation time, and the number of generations in which the optimal chromosome adaptation value does not change. During the selection operation, the chromosome with the highest fitness value in the population can always be copied. For the termination rule of the genetic algorithm, the chromosome with the largest adaptive value for 20 to 50 consecutive generations remains unchanged. When the termination condition is met, the chromosomes will be sorted according to the size of the adaptation value (dynamic integration scheme), and the output results will be used for decision-makers' reference.

## 5  Simulation Analysis of Evolution Process

Taking intercepting ballistic missiles as an example, we propose a strategic early warning concept [6]. The strategic early warning system mainly includes intelligence acquisition capabilities, intelligence processing capabilities, intelligence command coordination capabilities, integrated reconnaissance, surveillance, early warning capabilities, and intelligence transmission capabilities. The specific experimental scenario is that: after being upgraded, the system detection range is increased from 1000km to 7000km, the warning time is promoted from more than 1h to 1.5h, and the positioning accuracy is risen by 4%. The experimental given strategic early warning system has 22 optional systems. There are $2^{m(m+1)} = 2^{253}$ possible connection ways to realize 5 kinds of capabilities of the system. Based on the previous experience, the system designer selects 10 types of architecture construction schemes to form the initial architecture set $A_0$. The elements in $A_0$ are chromosomes $c_i$. Each chromosome $c_i$ represents an architecture scheme $a_i$. Through evolution, interfaces are continuously generated until a feasible solution, and eventually 100 chromosomes are generated, each of which is a recommended alternative architecture. According to the chromosome table, the average contribution of the system is 0.65, and the average value of a successful connection of the interface is 0.34. It indicates that in order to maintain a feasible chromosome, the actual connectivity is reduced. Table 2 shows the systems and feasible connections that each chromosome plays.

**Table 2: 10 examples of feasible chromosomes**

| number | chromosomes | System contribution | Interface contribution |
|---|---|---|---|
| 1 | 0110110011000000000001100110100011000000100110011000000 | 0.6 | 0.27 |
| 2 | 00111110100000000000000001010100111101111011001000010 | 0.7 | 0.33 |
| 3 | 10010110110010110100000000000000000011011000001001011001 | 0.6 | 0.29 |
| 4 | 0101011100000000000001111000000000111000000011101101000 | 0.6 | 0.29 |
| 5 | 10111100001111000000000001110000110000100000000000000 | 0.5 | 0.22 |
| 6 | 110011011110011001100110111000000000000010101010100011 1 | 0.7 | 0.40 |
| 7 | 110110110100001101011011010000000101101001010000101010 | 0.7 | 0.38 |
| 8 | 10111111010111111010000000011111011111011111001101101010 | 0.8 | 0.60 |
| 9 | 00100111100000000000000000000111100000000000101111011 1 | 0.6 | 0.27 |
| 10 | 001101111100000000000000000001011011110000000111111111 1 | .7 | 0.38 |

Further, to describe the initial state of each system, an excel table is provided for each system in each chromosome in Table 2. That is, an Excel data file is constructed for the 22 participating systems $S_i (i = 1,2,\cdots,22)$ in the strategic early warning system, which is used to describe the capabilities that each member system can provide and the connection relationship between the system and other systems. These excel data are used as the initial structure of the system, and Matlab is used to repeatedly weigh and calculate to find the best suitability value.

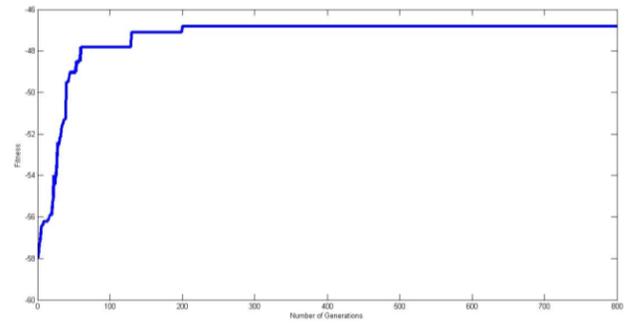

**Figure 1: MATLAB simulation of genetic algorithm**

The genetic sequence evolution in the genetic algorithm corresponds to the random change and evolution of the interface. Basically, the system evolution goes through five evolutionary transition states: preparation, initialization, collaboration, selection, and exit of member systems. In each state, it is necessary to negotiate with the system to see if the system agrees to cooperate. Given the state changes at each stage, each chromosome has evolved and output the results to arrive at an architecture integration scheme.

Based on this, we have developed an experimental platform for system evolution analysis. The initial combat mission of the simulation experiment is set through the platform, and the evaluation parameters (including architecture performance, flexibility, availability, and robustness) and experimental parameters (including combat mission attributes, system element attributes, etc.) are set to obtain the simulation experiment results. Assume that the cross probability of the poor individual is $P_{c1} = 0.9$, the cross probability of the best individual is $P_{c2} = 0.4$, the maximum probability of variation is $P_{m1} = 0.2$, the minimum probability of variation is $P_{m2} = 0.1$. The optimal system set and architecture after evolution are shown in Figure 2. The genetic algorithm adaptation value during the evolution process. The curve is shown in Figure 3.

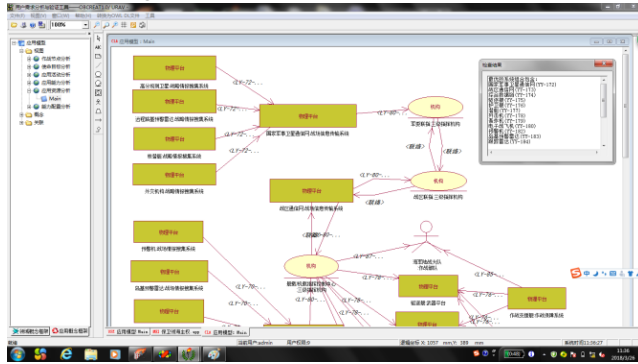

**Figure 2: The optimal system and architecture after evolution**

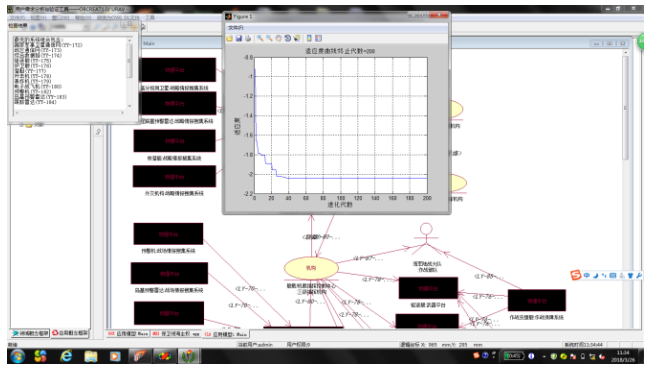

**Figure 3：Fitness value curve**

## 6   Summary

The network information system has the characteristic of system evolution. This paper describes each system and interface relationship of the network information system with a chromosome and then imports the chromosome data into an Excel database. From the perspective of biological genetic evolution, chromosomes are used as the specific encoding of genetic algorithms. We make use of the characteristics of chromosomal inheritance, crossover, mutation, as well as cooperative, speculative, and selfishness among the simulation systems to predict the evolution of the architecture. It provides a theoretical basis for the subsequent evolutionary evaluation analysis in the process of network information construction

## ACKNOWLEDGMENTS

This research was partially supported by grants from the National Natural Science Foundation of China （No.61802428, No.51874292） and China Postdoctoral Science Foundation of China (No. 2019M651991)

## REFERENCES

[1] Paulette Acheson, Louis Pape, Cihan Dagli, Nil Kilicay-Ergin, John Columbi, Khaled Haris. 2012. Understanding System of Systems Development Using an Agent- Based Wave Mode. *Procedia Computer Science*. Vol 12. 21-30. https://doi.org/10.1016/j.procs.2012.09.024.
[2] Dahmann J, Rebovich G, Lane J. A, Lowry R & Baldwin K. 2011. An Implementers' View of Systems Engineering for Systems Of Systems. In *Proceedings of IEEE International Systems Conference,* April 4-7, 2011, Montreal, QC, Canada, 212-217. https://doi.org/10.1109/SYSCON.2011.5929039.
[3] Ma Shaoping. 2004. *Artificial Intelligence.* Tsinghua University Press, Beijing.
[4] Su Qi,McAvoy Alex,Wang Long,Nowak Martin A. 2019. Evolutionary dynamics with game transitions. In *Proceedings of the National Academy of Sciences of the United States of America,* Vol 116, No.51. 25398-25404. https://doi.org/10.1073/pnas.1908936116.
[5] Deb K. & Goldberg D. E. 1989. An Investigation of Niche and Species Formation in Genetic Function Optimization. In *Proceedings of the Third International Conference on Genetic Algorithms.* June 1989. San Francisco, CA, United States.42-50.
[6] Zhang Tingting, Liu Xiaoming. 2016. Modeling and Evaluation of Emerging Capabilities in the Evolution of C4ISR System. *Military Operations Research and System Engineering.* Vol 30, No.1, 33-39.